\pdfoutput=1
\documentclass[letterpaper, 10 pt, conference]{ieeeconf}
\IEEEoverridecommandlockouts
\usepackage{graphics}
\usepackage{graphicx}
\usepackage{bm}
\usepackage{epsfig}
\usepackage{mathptmx}
\usepackage{times}
\usepackage[T1]{fontenc}
\usepackage{amsmath}
\usepackage{amssymb}
\usepackage{cite}
\usepackage{booktabs}
\usepackage{xcolor}
\usepackage{caption}
\usepackage{url}
\usepackage{hyperref}
\usepackage{tikz}
\usetikzlibrary{arrows.meta,positioning,fit,calc}

\newcommand{\GOLEM}{\textsc{Golem}}

\title{\LARGE \bf
Runtime Safety and Emergency Stops on a Full-Size Humanoid:\\Deployment Lessons and a Case for Graduated Force Authority
}

\author{
Max Conway$^{1*\dagger}$, Yutong Zhang$^{1*}$, Allen Devaraj$^1$, William Xie$^1$, \\
Alessandro Roncone$^1$, Nikolaus Correll$^{1,2}$
\thanks{$^1$Department of Computer Science, University of Colorado Boulder, USA.}
\thanks{$^2$Department of Aerospace and Mechanical Engineering, University of Notre Dame, USA.}
\thanks{$^*$Equal contribution. $^\dagger$Corresponding author.}
}

\begin{document}
\maketitle
\thispagestyle{empty}
\pagestyle{empty}

\begin{abstract}
Full-size humanoid robots are being considered for hazardous industrial work such as disassembling electric-vehicle batteries. We describe the safety mechanisms of an open-source autonomy stack for the Unitree H1-2 humanoid and what deploying them taught us. All joint commands pass through a safety layer that runs separately from the control software. It enforces joint limits and stops the robot when a state estimate becomes stale or the emergency stop (e-stop) is pressed or disconnected. We pushed the standing robot until it lost balance. With a model-predictive lower-body controller it never recovered, and the operator had to press the e-stop in every trial. Pressing it removes balance control, and the robot falls. In simulation, a controlled descent before the stop cut the impact energy tenfold, but only while the robot was still balanced. The safety layer's own limits trigger the same stop, and limits chosen for manipulation were exceeded during normal walking. Near an operator or a battery pack, the fall can be more dangerous than the cause of the stop. Joint limits bound how much force a robot can apply, but not when applying it is appropriate. We therefore make a case for granting robots force authority in stages, with a person or organization explicitly responsible for its use, work it through for battery screw extraction, and relate it to how people are granted dangerous capabilities.
\end{abstract}

\section{Introduction}

\begin{figure}[!t]
    \centering
    \includegraphics[width=0.66\linewidth]{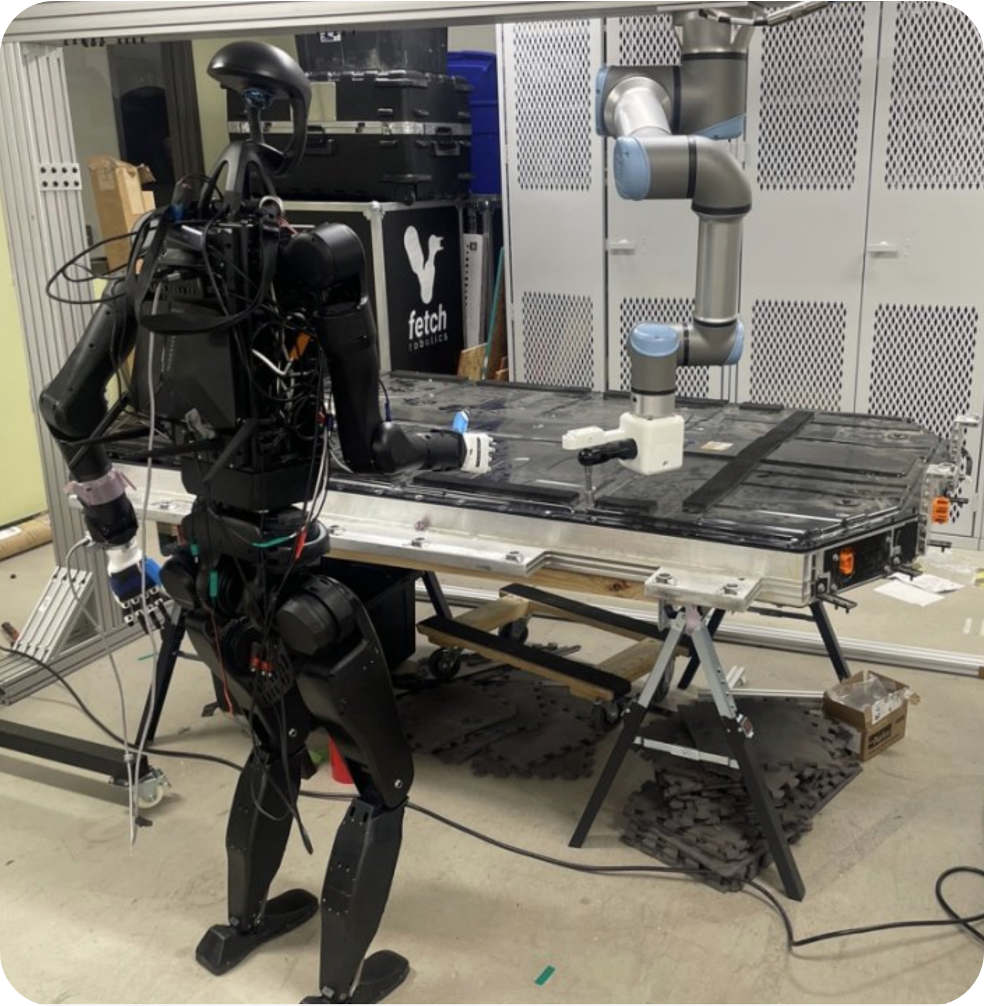}
    \caption{A Unitree H1-2 humanoid working alongside the gantry-based RAPID disassembly cell \cite{allen2026robotic} on a Hyundai Ioniq~5 battery pack.}
    \label{fig:teaser}
    \vspace{-14pt}
\end{figure}

End-of-life lithium-ion batteries from electric vehicles should be disassembled so that their materials can be recovered \cite{harper2019recycling}. The work is hazardous. Packs weigh about 450\,kg, carry up to 800\,V, arrive in an unknown state of charge, and can enter thermal runaway or electrocute workers \cite{allen2026robotic}. Gantry automation handles part of the task \cite{allen2026robotic,wegener2015robot}, but recycling facilities are built around human workers, so a full-size humanoid that fits into the existing cell without ramps or fixtures is attractive (Fig.~\ref{fig:teaser}). Such a humanoid is also a 70\,kg machine that must actively balance, and when it fails, it usually falls. Under geometric scaling, mass grows with the cube of height, and both the joint torque needed to support it and the fall energy grow with the fourth power. The 1.78\,m H1-2 has three times the knee torque of the 1.32\,m, 35\,kg Unitree G1 (360 vs.\ 120\,Nm) \cite{unitreeh12motors,unitreeg1}, close to the $(1.78/1.32)^4\approx3.3$ that this scaling predicts.

This extended abstract describes the safety-relevant parts of \GOLEM~(Generalized Open Library of Embodied Modules) \cite{conway2026golem}, an open-source ROS~2 stack for the H1-2 in which lower-body control, upper-body control, navigation, perception, and grasping are interchangeable modules behind fixed interfaces. Our contributions are:
\begin{enumerate}
    \item the \textbf{runtime safety layer} between the control modules and the motors, and deployment observations showing that its own limits can trigger a hazardous stop (Sec.~\ref{sec:safety});
    \item the \textbf{emergency stop}, push trials in which the robot lost balance and needed it, a simulation comparing a torque cut with a controlled descent, and an analysis of why stopping a humanoid is itself a hazard that industrial stop categories do not address (Sec.~\ref{sec:estop});
    \item a case for \textbf{graduated force authority}: fixed limits do not settle whether a force should be applied, so we argue that force envelopes be widened in stages and tied to a responsible party, with a worked example for screw extraction and precedents from developmental psychology and research on responsibility (Sec.~\ref{sec:force}).
\end{enumerate}

\section{System Context}\label{sec:system}

The architecture, controllers, and manipulation pipeline are described in \cite{conway2026golem}; here we summarize only what matters for safety. Each lower-body controller maps a commanded base velocity to leg joint commands. We compare two such controllers: MuJoCo-MPC (MJPC) \cite{howell2022predictive}, a sampling-based model predictive controller, and ALMI \cite{shi2025almi}, a learned policy trained adversarially against upper-body disturbances. We write $a^l$ and $a^u$ for the commanded lower- and upper-body joint positions. The H1-2's actuators deliver up to 360\,Nm at the knee and 120\,Nm at the shoulder and elbow \cite{unitreeh12motors}, and its grippers apply up to 32\,N \cite{correll2024versatile}.

\section{The Runtime Safety Layer}\label{sec:safety}

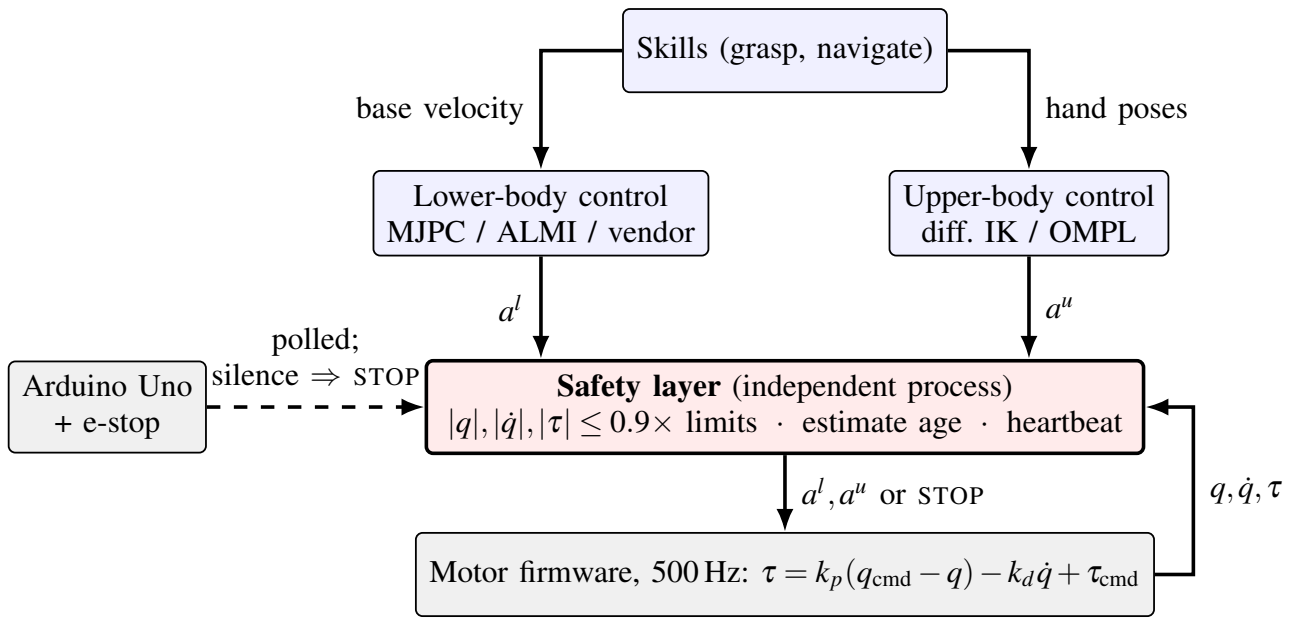
\begin{figure}[!t]
\centering
\resizebox{\linewidth}{!}{
\begin{tikzpicture}[
  font=\scriptsize,
  node distance=4mm and 4mm,
  box/.style={draw, rounded corners=1.5pt, align=center, inner sep=3pt, minimum height=6.5mm},
  mod/.style={box, fill=blue!6},
  saf/.style={box, fill=red!8, thick},
  hw/.style={box, fill=gray!12},
  arr/.style={-{Latex[length=1.8mm]}, thick},
  ]
  \node[mod] (loco) {Lower-body control\\MJPC / ALMI / vendor};
  \node[mod, right=14mm of loco] (ik) {Upper-body control\\diff.\ IK / OMPL};
  \path (loco.north) -- (ik.north) coordinate[midway] (midt);
  \path (loco.south) -- (ik.south) coordinate[midway] (midb);
  \node[mod, above=6mm of midt] (skill) {Skills (grasp, navigate)};
  \node[saf, below=8mm of midb, minimum width=56mm] (safety) {\textbf{Safety layer} (independent process)\\
    $|q|,|\dot q|,|\tau| \le 0.9\times$ limits $\;\cdot\;$ estimate age $\;\cdot\;$ heartbeat};
  \node[hw, below=6mm of safety, minimum width=56mm] (motor) {Motor firmware, 500\,Hz:\ $\tau = k_p(q_{\mathrm{cmd}}-q) - k_d\dot q + \tau_{\mathrm{cmd}}$};
  \node[hw, left=17mm of safety, align=center] (estop) {Arduino Uno\\+ e-stop};
  \draw[arr] (skill.west) -| (loco.north) node[pos=0.75, left] {base velocity};
  \draw[arr] (skill.east) -| (ik.north) node[pos=0.75, right] {hand poses};
  \draw[arr] (loco.south) -- (loco.south |- safety.north) node[midway, left] {$a^l$};
  \draw[arr] (ik.south) -- (ik.south |- safety.north) node[midway, right] {$a^u$};
  \draw[arr] (safety.south) -- (motor.north) node[midway, right] {$a^l, a^u$ or \textsc{stop}};
  \draw[arr] (motor.east) -- ++(3mm,0) |- (safety.east) node[pos=0.25, right] {$q,\dot q,\tau$};
  \draw[arr, dashed] (estop.east) -- (safety.west) node[midway, above, align=center, yshift=1pt] {polled;\\silence $\Rightarrow$ \textsc{stop}};
\end{tikzpicture}}
\caption{Every joint command passes through the safety layer, a process independent of the learned and model-based control modules. It enforces joint position, velocity, and torque limits at 90\% of the vendor specification, treats a stale state estimate or a lost heartbeat as a fault, and polls a microcontroller connected to a physical e-stop. Losing contact with the microcontroller also stops the robot.}
\label{fig:safety}
\vspace{-12pt}
\end{figure}

All motion commands $a^l, a^u$ pass through the safety layer (Fig.~\ref{fig:safety}), which runs as a separate low-level process. We made it independent because the modules upstream include a learned policy whose failure modes cannot be enumerated and a sampling-based optimizer whose output depends on an estimated base velocity. It continuously compares measured joint positions, velocities, and torques with configured limits and stops the robot as soon as any limit is exceeded, for example when a hand jams a screw against its bore, or when an upstream node stalls and the state estimate becomes stale. The limits are 90\% of the manufacturer's maximum values. This margin is a placeholder set by convention. A hazard analysis would derive each limit from what it protects: limits that bound contact with people from the body-region force and pressure thresholds of ISO/TS~15066 \cite{isots15066}, mapped through the arm's kinematics, and limits that protect the actuators from the vendor's thermal and current ratings. ISO/CD~25785-1 \cite{iso25785} could require such an analysis for robots that balance actively.

Two observations from deployment show that choosing the limits is non-trivial. First, no single, reliable set of limits for the H1-2 is available. The official walking policy shipped with the robot sometimes exceeds the velocity and torque limits in the released URDF model, so enforcing those limits as written would stop the robot during the vendor's own walking behavior. Second, we added a cap on elbow torque to limit how hard the arm can push on its surroundings. During walking, footstrikes produce torque spikes at the elbow that exceed this cap. The safety layer would then trigger the e-stop although the arm touches nothing, and torque would be switched off on all motors, including those in the legs, mid-stride.

\section{The Emergency Stop}\label{sec:estop}

\subsection{Implementation}
The safety layer continuously polls an Arduino Uno connected to a physical e-stop button, which lets a human supervisor halt the robot at any time. Losing contact with the Arduino is itself a stop condition. The e-stop is the only input to the safety layer that carries human judgment. Because the button is read by software, this e-stop lacks the integrity and the independence from the control software that an e-stop should have; a hard-wired safety relay would provide them. A relay that removes motor power immediately would make the Category~0 stop of Sec.~\ref{sec:notsafe} more reliable, and the robot would still fall.

\begin{figure*}[t]
    \centering
    \includegraphics[width=0.99\textwidth]{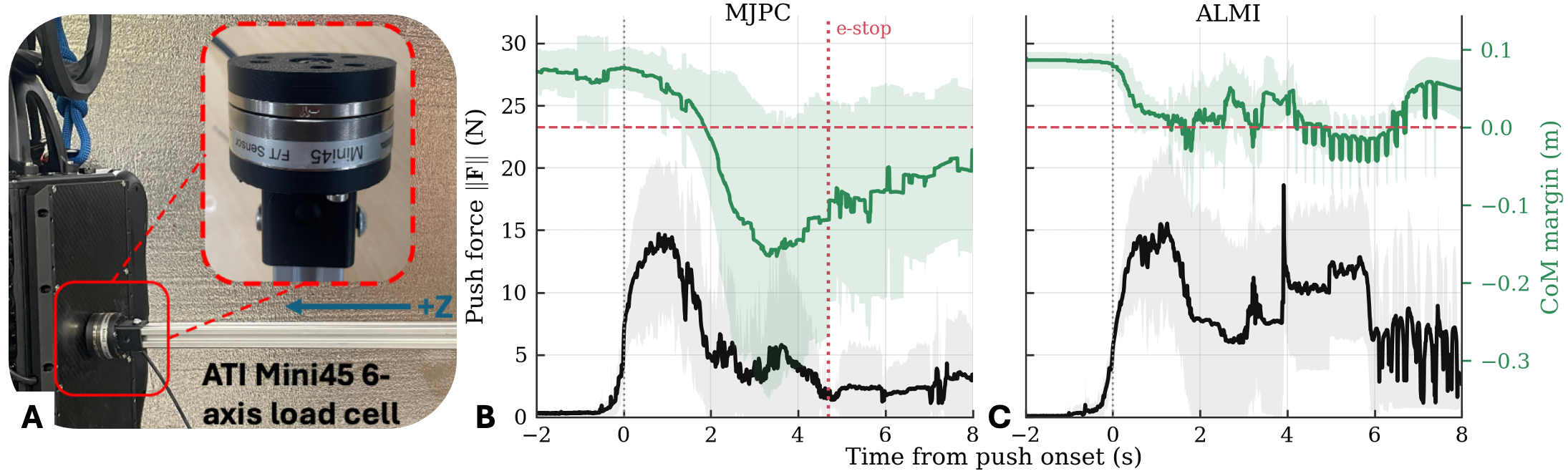}
    \caption{\textbf{(A)}~Pushes are applied with a rod instrumented with an ATI Mini45 six-axis load cell. \textbf{(B)}~Push force (\textbf{black}) and margin of the center of mass (CoM) to the support polygon (\textcolor{green!50!black}{green}; below the horizontal \textcolor{red}{red} line the CoM has left the polygon), mean and spread over 15 trials per controller. Left: under sampling-based MPC (MJPC), the margin decreases steadily after the push and the operator presses the e-stop (vertical \textcolor{red}{red} line) at $4.68\pm2.86$\,s. Right: the adversarially trained ALMI policy recovers and restores a positive margin in 14 of 15 trials without intervention.}
    \label{fig:pokey_data}
    \vspace{-14pt}
\end{figure*}

\subsection{When it was pressed}
Following \cite{conway2026golem}, we pushed the standing robot slightly above the center of the torso until it lost balance, that is, until its CoM left the support polygon (Fig.~\ref{fig:pokey_data}). The vendor's policy could not stand still and was excluded \cite{conway2026golem}. Both controllers run behind the same interface and the same safety layer, so the trials isolate the lower-body controller. Mean peak push force was similar for MJPC ($20.4\pm7.0$\,N) and ALMI ($22.1\pm6.9$\,N). Under MJPC, the CoM margin decreased steadily to $-0.166$\,m with a peak pitch lean of $16.3\pm6.3^\circ$, and the operator pressed the e-stop in every trial, at $4.68\pm2.86$\,s. ALMI reached a minimum margin of $-0.044$\,m with a peak lean of $6.8\pm1.4^\circ$ and recovered a positive margin in 14 of 15 trials. The e-stop decision was made by a person watching the robot lean, with a reaction time of seconds, and it worked only because the failure was slow.

\subsection{Stopping a humanoid is not a safe state}\label{sec:notsafe}
Industrial robot safety assumes that removing energy leads to a safe state. IEC~60204-1 \cite{iec60204} defines stop categories~0 (remove power immediately), 1 (controlled stop, then remove power), and 2 (stop under power), and ISO~13850 \cite{iso13850} requires an e-stop to use Category~0 or~1. The assumption holds for a fixed-base arm, whose brakes engage when power is removed, and for a statically stable wheeled manipulator.

It does not hold for a humanoid. The H1-2 stays upright only because a controller running at 50--67\,Hz keeps it there, and a Category~0 stop removes that controller. Our e-stop is a Category~0 stop that switches motor torque off. We chose it because it is the simplest stop the vendor firmware offers and the only one we could trust, since it does not depend on any controller. The robot, whose CoM is about 0.9\,m above the floor, then collapses. Its potential energy, $mgh\approx 70\,\mathrm{kg}\times9.81\,\mathrm{m/s^2}\times0.9\,\mathrm{m}\approx600$\,J, is released into the floor and whatever is in the way: a knee, a hand, the head-mounted LiDAR, or the operator who pressed the button and is standing nearby. Falling onto the exposed cells of a pack could start the thermal runaway that robotic disassembly is meant to keep away from people. In our push trials, an overhead safety line caught the robot, so we have not measured a fall.

Cutting the controller's authority removes the hazard from the controller but creates one from the body. A benign stop would require a controlled descent to a kneel or a sit, or a damage-minimizing fall \cite{fujiwara2002ukemi,ha2015fall}. These are whole-body balance behaviors of exactly the kind the safety layer is meant to be independent of.

\begin{figure}[t]
\centering
\includegraphics[width=\linewidth]{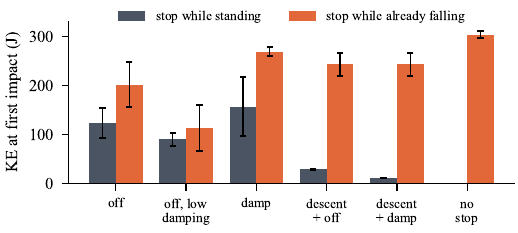}
\caption{Simulated H1-2 (MuJoCo digital twin \cite{conway2026golem}): kinetic energy at the first floor contact of a body other than a foot, after a stop requested while standing or after an unrecoverable push (mean, s.d., 10 runs). \emph{Off}: torque cut, with the modeled passive joint damping or one tenth of it; \emph{damp}: $\tau=-20\,\dot q$; \emph{descent}: 1.5\,s to a crouch, then off or damping. Holding the crouch never fell.}
\label{fig:falls}
\vspace{-12pt}
\end{figure}

To see how much a descent helps, we simulated the stops on the H1-2 digital twin (Fig.~\ref{fig:falls}). A torque cut from standing released $508\pm25$\,J, but only $124\pm31$\,J was kinetic energy at the first impact, usually of the torso; joint friction and later contacts absorbed the rest. A 1.5\,s descent to a crouch lowered the CoM from 0.92 to 0.56\,m without a fall, and damping from the crouch reduced the impact to 11\,J. After a push the robot could not recover from, the descent, a joint trajectory without balance feedback, did no better than a torque cut ($243\pm23$ vs.\ $202\pm45$\,J). Damping alone was worse than a torque cut in both cases, because the robot toppled without folding at the knees. Our trials sharpen the problem: the only controller in our stack that recovers is the learned one, so switching to ``the simple verified controller'' on e-stop would switch away from recovery. Recent work learns whether a safe stop is still feasible from the current state \cite{sun2026stoppability}, and executes a learned stop policy only when it is, falling back to damping otherwise \cite{long2026safestop}. In that simulation study on a smaller humanoid, always damping produced catastrophic contacts in 5.4\% of episodes, compared with 1.4--1.7\% when a stop was attempted. Both damping and a learned stop keep the motors powered while the robot comes to rest, which makes them Category~1 stops at best, and the learned stop depends on a whole-body policy working during the emergency. These methods decide how to stop once a stop is requested. In our stack, the safety layer requests stops itself, so a false limit trip, such as the elbow cap during walking, becomes a fall. Limits and stop behavior therefore cannot be designed separately.

We see three partial answers. (1)~A \emph{stop hierarchy}: a freeze that keeps the lower-body controller running with a zero-velocity command when the hazard is the arm, which is the common default but works only within the recoverable range \cite{long2026safestop}; a controlled descent when the hazard is the base; and damping as the last level, with a power cut only when the robot is restrained or on the ground. ISO/CD~25785-1 \cite{iso25785}, under development for industrial mobile robots with actively controlled stability, is the natural place to define such a hierarchy. (2)~A \emph{minimal, verifiable balance fallback}, for example control barrier functions \cite{ames2019cbf} on CoM margin and joint torque around a crouch-and-kneel primitive, small enough to analyze, so that safety does not rely on the learned policy's recovery. (3)~Until these exist, \emph{physical restraint} by tether, hoist, or fence, with e-stop reaction times measured and reported.

\section{A Case for Graduated Force Authority}\label{sec:force}

\subsection{Fixed limits do not encode context}
The elbow cap of Sec.~\ref{sec:safety} is the simplest case: a value that limits how hard the arm pushes during manipulation is exceeded by footstrikes during walking, so no single value serves both. In our screw-removal trials, all five failures began with off-axis extraction, after which the grasp force was \emph{too low} to overcome the resulting friction \cite{conway2026golem}, yet the force that frees a stuck screw could injure a wrist. ISO/TS~15066 permits different contact forces for different body regions \cite{isots15066}, and injury tolerance depends on the body part and the type of contact \cite{haddadin2009requirements}, but neither a torque limit nor a body-region table encodes \emph{why} a force is applied. Xie~et~al.~\cite{xie2025dualuse} found that vision-language models planning grasp forces on human body parts used the same physical reasoning for helpful and harmful requests. A simple safeguard reduced harmful grasps from 67\% to under 2\%, and helpful ones from 91\% to 45\%. A rule such as ``never exceed 32\,N'' likewise blocks both harm and screw removal.

Fallback rules have the same problem. In a 2023 crash, a Cruise robotaxi stopped with a pedestrian underneath it, classified the event as a side impact, and began a pullover to a ``minimal risk condition'' that dragged her 6\,m \cite{koopman2024anatomy}. Her legs were visible in a side camera but were not tracked. Like our e-stop, the designated safe response caused most of the harm. Avoiding it required object permanence, which infants reach within their first year \cite{piaget1954construction,baillargeon1985object}, and the judgment that a rule meant to reduce risk does not apply when following it increases risk.

\subsection{Graduated force envelopes}
We propose that the safety layer's limits become task- and context-dependent envelopes that start conservative, widen only for contact modes the system has shown it can recognize and handle, and return to the conservative default when the context no longer matches. Each widening is recorded with the person or organization responsible for the robot's use of force in that context. \GOLEM's evaluation is already a capability ladder \cite{conway2026golem}: the robot is first hoist-tethered, then free-standing, then standing after navigating to the pack, and grasp success falls from 97\% to 87\% to 37\% across these rungs.

For battery screw extraction, the envelope would work as follows. (i)~\emph{Default:} grasp force and arm torques stay at values acceptable for incidental contact with a person's hand. (ii)~\emph{Evidence for widening:} a record of extractions on the same pack model at the default level, with failures classified by cause, here off-axis extraction followed by insufficient grasp force; at runtime, a confident detection of the screw head on the pack and a pull direction, estimated from joint torques, aligned with the screw axis. (iii)~\emph{Approval:} the engineer responsible for the cell approves the wider envelope for this contact mode and pack model, the approval is logged with their name, and the supervisor holding the e-stop enables it for a shift. (iv)~\emph{Reversion:} the envelope returns to the default when a person enters the workspace, the detection confidence drops, the pull direction leaves the screw axis, which preceded all five failures in our trials, or the task changes. The elbow cap fits the same structure, with walking and manipulation as separate contact modes.

\subsection{Precedents from development and responsibility}
Robots do not develop as children do. People nonetheless face the same two questions with children: when to trust an agent with a dangerous capability, and who answers for its actions until then. Piaget distinguished a \emph{heteronomous} morality of rules obeyed as given from an \emph{autonomous} one that weighs intentions, which children reach around age nine \cite[pp.~120--129]{piaget1932moral}. A torque limit, the safeguard in \cite{xie2025dualuse}, and the robotaxi's pullover rule are heteronomous. Graduated driver licensing \cite{shope2007graduated}, scaffolding \cite{wood1976tutoring,vandepol2010scaffolding}, and entrustable professional activities in medical training \cite{tencate2005entrustability,tencate2013nuts} grant dangerous capabilities in stages tied to demonstrated competence and supervision. People are held responsible to the degree that they caused an outcome and could control it \cite{weiner1995judgments,malle2014theory}, and a learning system can act in ways no one can predict, which leaves a gap in responsibility \cite{matthias2004responsibility}. Naming a responsible party for each widening is meant to close that gap for force, with reliance calibrated to the robot's actual capabilities \cite{lee2004trust}.

\section{Discussion and Limitations}\label{sec:discussion}

Our evidence is narrow. It comes from one robot and from push trials while standing, with 15 trials per controller. A safety line caught every fall, so our fall energies come from a simulation with simplified contacts, in which the robot's links do not collide with each other, and the passive damping of unpowered joints is unknown; controlled drops onto a padded, instrumented plate would measure them. We did not log how often the elbow cap was exceeded while walking; a trip rate per minute of walking at a given speed would show how often a fixed limit stops the robot. None of the three answers to the stop problem is implemented on hardware. Choosing among the levels of a stop hierarchy at runtime requires knowing whether the robot can still recover \cite{sun2026stoppability,long2026safestop}, which our simulated descent ignores. The digital twins in \cite{conway2026golem} expose the same ROS~2 interfaces as the hardware, so our safety layer and stop logic could be compared with these methods in simulation first, without repeatedly dropping a 70\,kg robot.

Graduated force authority has limitations of its own. A robot's competence changes abruptly when a model is retrained or a controller is swapped, which may invalidate authority granted earlier, and benchmarks for a robot's judgment about force do not yet exist. Context-dependent envelopes widen when perception reports high confidence, so their safety depends on the calibration of a learned perception stack, the kind of component the safety layer was designed not to trust. Finally, responsibility is already divided in our own system. The vendor's walking policy and its released limits disagree, and our safety layer sits between them. Assigning responsibility for force will require deciding who answers for the integrated system, not only for each component.

\section{Conclusion}
Deploying a safety layer and e-stop on a full-size humanoid taught us four things. First, on a robot that must actively balance, the e-stop does not lead to a safe state: it removes balance control, and the robot falls; in simulation, a controlled descent cut the impact energy tenfold, but only while the robot was still balanced. Second, choosing joint limits is non-trivial, and because limit violations trigger the same stop, limits and stop behavior have to be designed together. Third, the controller that recovered from pushes was the learned one, which we cannot certify. Fourth, force limits cannot tell when force is appropriate, so force authority should be granted in stages, with an explicitly responsible party.

\section*{Acknowledgment}
The authors are supported by ARPA-E grant DE-AR0001966, ``Robust Robotic Disassembly of EV Battery Packs using Open-World Vision Language Models and Symbolic Replanning.'' William Xie is supported by the NSF Graduate Research Fellowship. N.~Correll and M.~Conway have an interest in Realtime Manufacturing, Inc., which is working on humanoids for manufacturing applications.

\bibliographystyle{ieeetr}
\bibliography{HAMS}

\begin{thebibliography}{10}

\bibitem{harper2019recycling}
G.~Harper, R.~Sommerville, E.~Kendrick, L.~Driscoll, P.~Slater, R.~Stolkin,
  A.~Walton, P.~Christensen, O.~Heidrich, S.~Lambert, {\em et~al.}, ``Recycling
  lithium-ion batteries from electric vehicles,'' {\em Nature}, vol.~575,
  no.~7781, pp.~75--86, 2019.

\bibitem{allen2026robotic}
Z.~Allen, M.~Conway, L.~Antieau, A.~Devaraj, and N.~Correll, ``Robotic agentic
  platform for intelligent electric vehicle disassembly,'' in {\em IEEE 22nd
  International Conference on Automation Science and Engineering (CASE)},
  (Shenyang, China), pp.~966--973, 2026.

\bibitem{wegener2015robot}
K.~Wegener, W.~H. Chen, F.~Dietrich, K.~Dr{\"o}der, and S.~Kara, ``Robot
  assisted disassembly for the recycling of electric vehicle batteries,'' {\em
  Procedia CIRP}, vol.~29, pp.~716--721, 2015.

\bibitem{unitreeh12motors}
{Unitree Robotics}, ``{Unitree H1} and {H1-2} product specifications.''
  \url{https://www.unitree.com/h1}.
\newblock Accessed October 2026.

\bibitem{unitreeg1}
{Unitree Robotics}, ``{Unitree G1} product specifications.''
  \url{https://www.unitree.com/g1}.
\newblock Accessed November 2026.

\bibitem{conway2026golem}
M.~Conway, W.~Xie, A.~Devaraj, Y.~Zhang, N.~Pudasaini, M.~Feit, A.~Abid,
  Z.~Allen, C.~Liu, X.~Tan, J.~Lavering, J.~Chen, L.~Antieau, A.~Von~Pischke,
  A.~Roncone, Z.~Sunberg, and N.~Correll, ``{GOLEM}: Modular humanoid autonomy
  towards electric vehicle battery disassembly,'' {\em arXiv preprint
  arXiv:2608.21550}, 2026.

\bibitem{howell2022predictive}
T.~Howell, N.~Gileadi, S.~Tunyasuvunakool, K.~Zakka, T.~Erez, and Y.~Tassa,
  ``Predictive sampling: Real-time behaviour synthesis with {MuJoCo},'' {\em
  arXiv preprint arXiv:2212.00541}, 2022.

\bibitem{shi2025almi}
J.~Shi, X.~Liu, D.~Wang, O.~Lu, S.~Schwertfeger, C.~Zhang, F.~Sun, C.~Bai, and
  X.~Li, ``Adversarial locomotion and motion imitation for humanoid policy
  learning,'' in {\em Advances in Neural Information Processing Systems},
  vol.~38, pp.~73918--73949, 2025.

\bibitem{correll2024versatile}
N.~Correll, D.~Kriegman, S.~Otto, and J.~Watson, ``A versatile robotic hand
  with {3D} perception, force sensing for autonomous manipulation,'' {\em arXiv
  preprint arXiv:2402.06018}, 2024.

\bibitem{isots15066}
{International Organization for Standardization}, ``{ISO/TS 15066:2016 Robots
  and robotic devices --- Collaborative robots}.'' Geneva, Switzerland, 2016.

\bibitem{iso25785}
{International Organization for Standardization}, ``{ISO/CD 25785-1 Robotics
  --- Safety requirements for dynamically stable industrial mobile robots
  (legged, wheeled, or other forms of locomotion) --- Part 1: Robots}.''
  Committee Draft, stage 30.60, ISO/TC 299, 2026.

\bibitem{iec60204}
{International Electrotechnical Commission}, ``{IEC 60204-1:2016+AMD1:2021
  Safety of machinery --- Electrical equipment of machines --- Part 1: General
  requirements}.'' Geneva, Switzerland, 2021.

\bibitem{iso13850}
{International Organization for Standardization}, ``{ISO 13850:2015 Safety of
  machinery --- Emergency stop function --- Principles for design}.'' Geneva,
  Switzerland, 2015.

\bibitem{fujiwara2002ukemi}
K.~Fujiwara, F.~Kanehiro, S.~Kajita, K.~Kaneko, K.~Yokoi, and H.~Hirukawa,
  ``{UKEMI}: Falling motion control to minimize damage to biped humanoid
  robot,'' in {\em IEEE/RSJ International Conference on Intelligent Robots and
  Systems (IROS)}, vol.~3, pp.~2521--2526, 2002.

\bibitem{ha2015fall}
S.~Ha and C.~K. Liu, ``Multiple contact planning for minimizing damage of
  humanoid falls,'' in {\em IEEE/RSJ International Conference on Intelligent
  Robots and Systems (IROS)}, pp.~2761--2767, 2015.

\bibitem{sun2026stoppability}
Y.~Sun, Y.~Pan, S.~Li, C.~Ding, T.~Cui, L.~Wang, and C.~Liu, ``Learning
  safe-stoppability monitors for humanoid robots,'' {\em arXiv preprint
  arXiv:2603.22703}, 2026.

\bibitem{long2026safestop}
J.~Long, P.~Abbeel, K.~Sreenath, R.~Horowitz, G.~Shi, and C.~K. Liu, ``Humanoid
  safe stop via learned stoppability value,'' {\em arXiv preprint
  arXiv:2609.02358}, 2026.

\bibitem{ames2019cbf}
A.~D. Ames, S.~Coogan, M.~Egerstedt, G.~Notomista, K.~Sreenath, and P.~Tabuada,
  ``Control barrier functions: Theory and applications,'' in {\em 18th European
  Control Conference (ECC)}, pp.~3420--3431, 2019.

\bibitem{haddadin2009requirements}
S.~Haddadin, A.~Albu-Sch{\"a}ffer, and G.~Hirzinger, ``Requirements for safe
  robots: Measurements, analysis and new insights,'' {\em The International
  Journal of Robotics Research}, vol.~28, no.~11-12, pp.~1507--1527, 2009.

\bibitem{xie2025dualuse}
W.~Xie, E.~Rice, and N.~Correll, ``On the dual-use dilemma in physical
  reasoning and force,'' in {\em RSS 2025 Workshop on Reliable Robotics: Safety
  and Security in the Face of GenAI}, (Los Angeles, CA), 2025.
\newblock arXiv:2505.18792.

\bibitem{koopman2024anatomy}
P.~Koopman, ``Anatomy of a robotaxi crash: Lessons from the {C}ruise pedestrian
  dragging mishap,'' in {\em Computer Safety, Reliability, and Security
  (SAFECOMP)}, vol.~14988 of {\em LNCS}, pp.~119--133, Springer, 2024.

\bibitem{piaget1954construction}
J.~Piaget, {\em The Construction of Reality in the Child}.
\newblock New York: Basic Books, 1954.

\bibitem{baillargeon1985object}
R.~Baillargeon, E.~S. Spelke, and S.~Wasserman, ``Object permanence in
  five-month-old infants,'' {\em Cognition}, vol.~20, no.~3, pp.~191--208,
  1985.

\bibitem{piaget1932moral}
J.~Piaget, {\em The Moral Judgment of the Child}.
\newblock London: Kegan Paul, Trench, Trubner \& Co., 1932.
\newblock Translated by M. Gabain.

\bibitem{shope2007graduated}
J.~T. Shope, ``Graduated driver licensing: Review of evaluation results since
  2002,'' {\em Journal of Safety Research}, vol.~38, no.~2, pp.~165--175, 2007.

\bibitem{wood1976tutoring}
D.~Wood, J.~S. Bruner, and G.~Ross, ``The role of tutoring in problem
  solving,'' {\em Journal of Child Psychology and Psychiatry}, vol.~17, no.~2,
  pp.~89--100, 1976.

\bibitem{vandepol2010scaffolding}
J.~van~de Pol, M.~Volman, and J.~Beishuizen, ``Scaffolding in teacher--student
  interaction: A decade of research,'' {\em Educational Psychology Review},
  vol.~22, no.~3, pp.~271--296, 2010.

\bibitem{tencate2005entrustability}
O.~ten Cate, ``Entrustability of professional activities and competency-based
  training,'' {\em Medical Education}, vol.~39, no.~12, pp.~1176--1177, 2005.

\bibitem{tencate2013nuts}
O.~ten Cate, ``Nuts and bolts of entrustable professional activities,'' {\em
  Journal of Graduate Medical Education}, vol.~5, no.~1, pp.~157--158, 2013.

\bibitem{weiner1995judgments}
B.~Weiner, {\em Judgments of Responsibility: A Foundation for a Theory of
  Social Conduct}.
\newblock New York: Guilford Press, 1995.

\bibitem{malle2014theory}
B.~F. Malle, S.~Guglielmo, and A.~E. Monroe, ``A theory of blame,'' {\em
  Psychological Inquiry}, vol.~25, no.~2, pp.~147--186, 2014.

\bibitem{matthias2004responsibility}
A.~Matthias, ``The responsibility gap: Ascribing responsibility for the actions
  of learning automata,'' {\em Ethics and Information Technology}, vol.~6,
  no.~3, pp.~175--183, 2004.

\bibitem{lee2004trust}
J.~D. Lee and K.~A. See, ``Trust in automation: Designing for appropriate
  reliance,'' {\em Human Factors}, vol.~46, no.~1, pp.~50--80, 2004.

\end{thebibliography}

\end{document}